\documentclass[10pt,twocolumn,letterpaper]{article}

\usepackage[pagenumbers]{iccv} 

\usepackage{adjustbox}
\usepackage[table ]{xcolor}

\definecolor{lightblue}{RGB}{173,216,230} 
\definecolor{darkblue}{RGB}{0,0,139} 
\usepackage[utf8]{inputenc}

\definecolor{iccvblue}{rgb}{0.21,0.49,0.74}

\usepackage[pagebackref,breaklinks,colorlinks,allcolors=iccvblue]{hyperref}
\usepackage{afterpage}

\def\paperID{10178} 
\def\confName{ICCV}
\def\confYear{2025}

\title{\includegraphics[width=0.7cm]{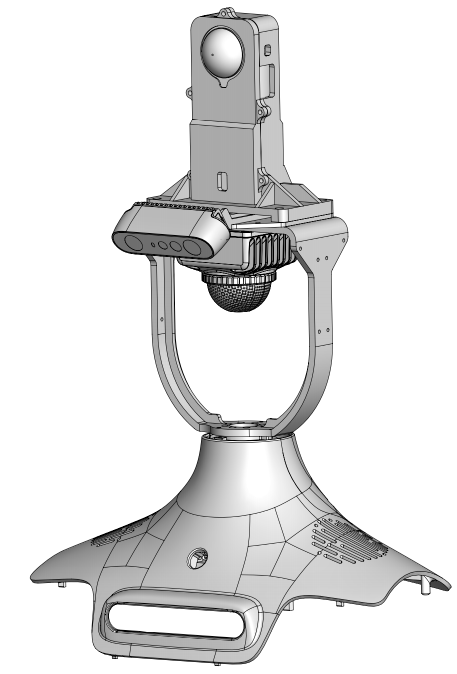}\textbf{\textcolor{iccvblue}{HumanoidPano}}: Hybrid Spherical Panoramic-LiDAR Cross-Modal Perception for Humanoid Robots}

\author{
    Qiang Zhang$^{1,2,*}$,
    Zhang Zhang$^{1,4,*}$,
    Wei Cui$^{1,*}$,
    Jingkai Sun$^{1,2}$,\\
    Jiahang Cao$^{1,2}$,
    Yijie Guo$^{1}$,
    Gang Han$^{1}$,
    Wen Zhao$^{1}$,
    Jiaxu Wang$^{1,2,3}$,\\
    Chenghao Sun$^{1}$,
    Lingfeng Zhang$^{2}$,
    Hao Cheng$^{2}$,
    Yujie Chen$^{5}$,
    Lin Wang$^{6}$,\\
    Jian Tang$^{1,\dagger}$,
    Renjing Xu$^{2,\dagger}$\\
    {\tt\small jony.zhang@x-humanoid.com}\\
    $^{1}$ Beijing Innovation Center of Humanoid Robotics \\
    $^{2}$ Hong Kong University of Science and Technology (Guangzhou) \\
    $^{3}$ Hong Kong University of Science and Technology \\
    $^{4}$ Beijing Institute of Technology \\
    $^{5}$ ETH Zurich \\
    $^{6}$ Nanyang Technological University \\
    \footnotesize{$^*$ Contributed equally.}
    \footnotesize{$^\dagger$ Corresponding authors.}
}

\begin{document}
\twocolumn[{
\renewcommand\twocolumn[1][]{#1}
\maketitle
\begin{center}
    \captionsetup{type=figure}
    \includegraphics[width=2.0\columnwidth]{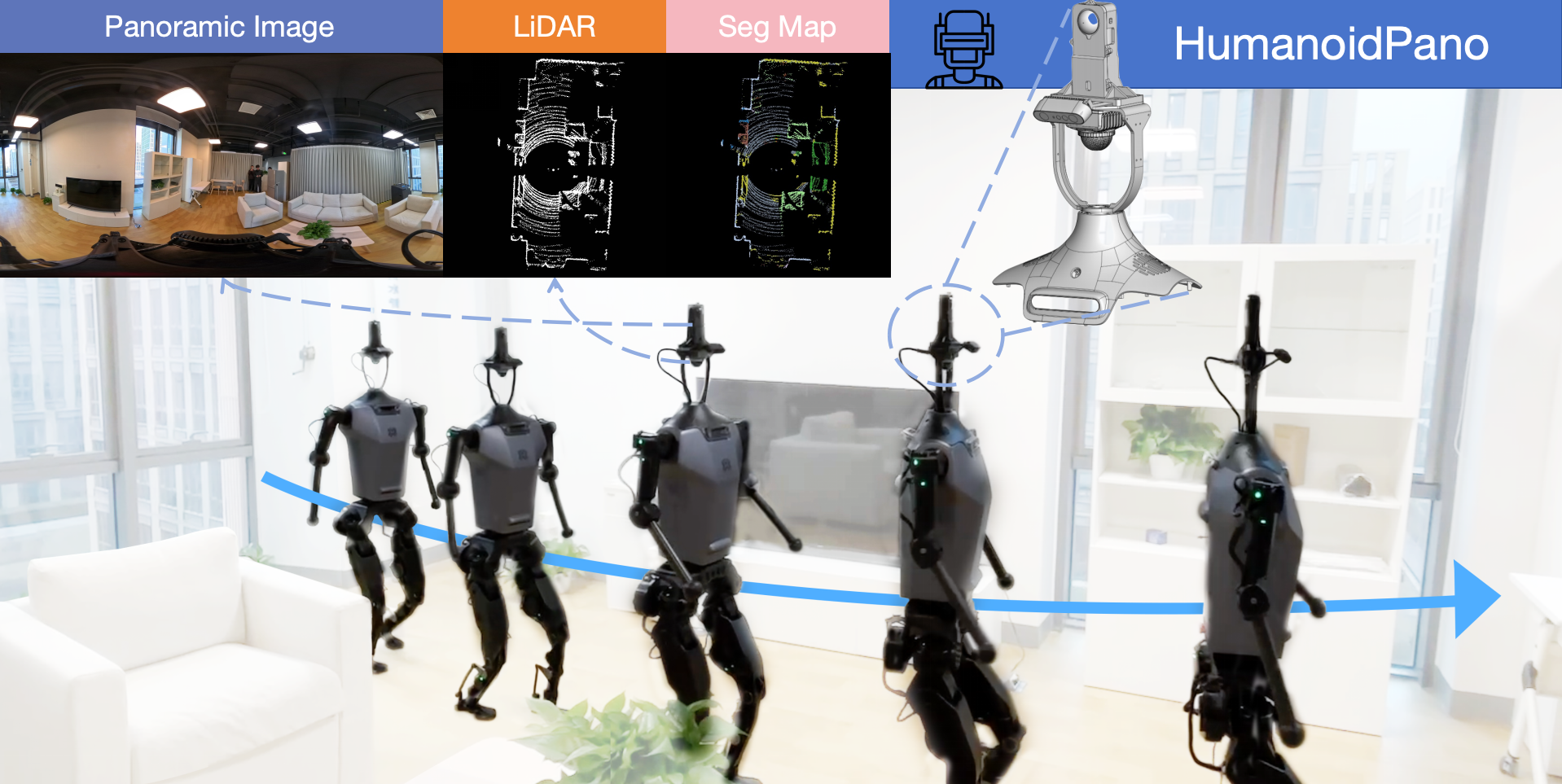}
    \captionof{figure}{\textbf{The humanoid robot autonomously navigates complex environments using \textcolor{iccvblue}{HumanoidPano}}, which fuses panoramic vision and LiDAR to generate real-time BEV semantic maps. The figure depicts motion trajectory while visualizing real-time visual sensor data and perception results in the upper-left corner, alongside detailed illustrations of the HumanoidPano universal perception module.}
    \label{walk}
\end{center}
}]

\begin{abstract}

The perceptual system design for humanoid robots poses unique challenges due to inherent structural constraints that cause severe self-occlusion and limited field-of-view (FOV). We present \textbf{\textcolor{iccvblue}{HumanoidPano}}, a novel hybrid cross-modal perception framework that synergistically integrates panoramic vision and LiDAR sensing to overcome these limitations. Unlike conventional robot perception systems that rely on monocular cameras or standard multi-sensor configurations, our method establishes geometrically-aware modality alignment through a spherical vision transformer, enabling seamless fusion of 360° visual context with LiDAR's precise depth measurements. 
First, Spherical Geometry-aware Constraints (SGC) leverage panoramic camera ray properties to guide distortion-regularized sampling offsets for geometric alignment. Second, Spatial Deformable Attention (SDA) aggregates hierarchical 3D features via spherical offsets, enabling efficient 360°-to-BEV fusion with geometrically complete object representations. Third, Panoramic Augmentation (AUG) combines cross-view transformations and semantic alignment to enhance BEV-panoramic feature consistency during data augmentation.
Extensive evaluations demonstrate \textbf{state-of-the-art} performance on the 360BEV-Matterport benchmark.
Real-world deployment on humanoid platforms validates the system's capability to generate accurate BEV segmentation maps through panoramic-LiDAR co-perception, directly enabling downstream navigation tasks in complex environments. Our work establishes a new paradigm for embodied perception in humanoid robotics.
The codes will be released soon.
\end{abstract}   
\section{Introduction}
\label{sec:intro}

Robotic environmental perception driven by computer vision has achieved revolutionary advancements\cite{levine2016end, kim2024openvla, wu2024robomind, brohan2022rt, zhang2025mapnav, chi2023diffusion, fu2024mobile, zhao2023learning, duan2024enhancing, qi2025sofar}, with groundbreaking innovations profoundly reshaping technological landscapes in core applications such as autonomous driving and industrial automation. Current industry standard solutions predominantly rely on rigid, occlusion free sensor configuration assumptions: Autonomous vehicles acquire unobstructed omnidirectional perception through rigid roof mounted multi-sensor arrays (LiDAR, camere, millimeter wave radar etc.)\cite{geiger2013vision, liao2022kitti, geiger2012we, caesar2020nuscenes, sun2020scalability, chang2017matterport3d, armeni2017joint}, while industrial robotic arms achieve exceptional positioning accuracy via fixed spatial relationships in Eye-in-Hand vision systems\cite{ha2024umi, chi2024universal}. These domains share a fundamental design paradigm ensuring sensor field-of-view (FOV) invariance through mechanical engineering. For instance, autonomous vehicles maintain inter sensor geometric consistency through rigid chassis structures, and warehouse robots eliminate dynamic occlusions using fixed height sensor columns.

Humanoid robots, widely regarded as the most promising candidate for general purpose robotics and embodying humanity's vision for the pinnacle of robotic embodiment, have recently garnered unprecedented attention\cite{zhang2024whole, fu2024humanplus, lu2024mobile, duan2024e2h, chen2025hifar, cuiadapting}. However, their environmental perception capabilities have progressed relatively slowly compared to rapid advancements in mechanical design and locomotion strategies\cite{yuan2023hierarchical}. The autonomous operation of humanoid robots in unstructured environments remains fundamentally constrained by an inherent paradox between their biomimetic structure and perceptual requirements. While anthropomorphic limb configurations enhance mobility, they simultaneously create persistent self-occlusion zones elbow joints obstruct chest mounted cameras during manipulation tasks, while thigh movements completely blind waist mounted sensors when navigating obstacles. This structure induced perceptual degradation renders conventional stereo vision solutions, widely adopted in general robotics, significantly less effective on humanoid platforms.

As previously analyzed, the development of visual perception systems for humanoid robots faces multifaceted challenges. While multiview sensing configurations have demonstrated efficacy in autonomous driving, their adaptation to humanoid platforms encounters dual constraints: severe data scarcity hinders data driven model adaptation, and morphological diversity prevents the establishment of universal sensor placement standards. Although panoramic vision overcomes field-of-view limitations, it introduces triple feature degradation under spherical projection—structural fragmentation of linear elements; cross-regional scale disparity; and depth-aware feature degradation. Existing decoders relying on 2D offset prediction mechanisms fail to model the coupled effects of distortion and scene depth. In multimodal fusion, automotive LiDAR-camera registration paradigms prove inadequate due to geometric incompatibility between spherical projection and point clouds, while dynamic joint motions induce real-time drift in cross modal geometric alignment matrices. Furthermore, the complex spatial mapping required for unified panoramic-BEV representations suffers from geometric paucity in training data, with conventional augmentation strategies unable to preserve cross view metric consistency—critical limitations that severely constrain 3D environmental comprehension.

We present HumanoidPano, a novel perception framework specifically designed for humanoid robots, achieving comprehensive environmental awareness through synergistic fusion of panoramic vision and LiDAR sensing. The framework operates through three stage hierarchical processing: pixel level semantic feature extraction via panoramic imaging, geometry-aware depth profiling using LiDAR point clouds, and cross modal feature fusion through our proposed geometrically grounded alignment mechanism. 

Our methodology reimagines robotic perception through four pivotal innovations:
\begin{itemize}
    \item \textbf{Spherical Geometry-aware Constraints. (SGC)}: Introduces geometric constraints derived from panoramic imaging principles, enabling distortion-aware cross-modal alignment through ray-based feature modulation.
    \item \textbf{Spatial Deformable Attention. (SDA)}: Achieves efficient 360°-to-BEV fusion via adaptive spatial deformable attention and sampling.
    \item \textbf{Panoramic Augmentation (AUG)}: Introduces a novel data augmentation approach that integrates panoramic image and BEV space transformations—including random flipping, mixing, and semantic map alignment—to address alignment challenges in joint view augmentation.
\end{itemize}
This geometrically-coherent framework establishes new standards for humanoid perception, effectively bridging the gap between panoramic distortion resilience and 3D scene understanding while maintaining real time operational efficiency.

We establishes new methodological standards for humanoid robotic perception. By transforming structural constraints into design assets, HumanoidPano demonstrates that humanoid robots can transcend biological vision limitations when perception algorithms are deeply adapted to their morphological characteristics. Our contributions bridge critical technical gaps in environmental comprehension for humanoid platforms while providing a scalable architectural blueprint for future embodied intelligence systems. The principal advancements are:

\begin{itemize}
    \item \textbf{A Universal Multimodal Humanoid Robots Perception Framework}: 
        \textbf{HumanoidPano}, a general purpose perception system optimized for humanoid biomechanical constraints, integrating panoramic vision and LiDAR through structural-aware fusion.
    \item \textbf{State-of-the-art} results on the 360BEV Matterport dataset, surpassing previous methods in the precision of understanding the panoramic BEV scene.
    \item \textbf{Real World Humanoid Robot Validation}: 
        Demonstrated practical perception capabilities on physical humanoid platforms, enabling advanced downstream tasks like semantic map based navigation in complex environments. (Figure \ref{walk})
\end{itemize}

These achievements collectively redefine the paradigm of environment interaction for humanoid robotics, establishing foundational infrastructure for next generation embodied AI systems.

\begin{figure*}[t]
	\setlength{\tabcolsep}{1.0pt}
	\centering
	\begin{tabular}{c}
  \includegraphics[width=\textwidth]{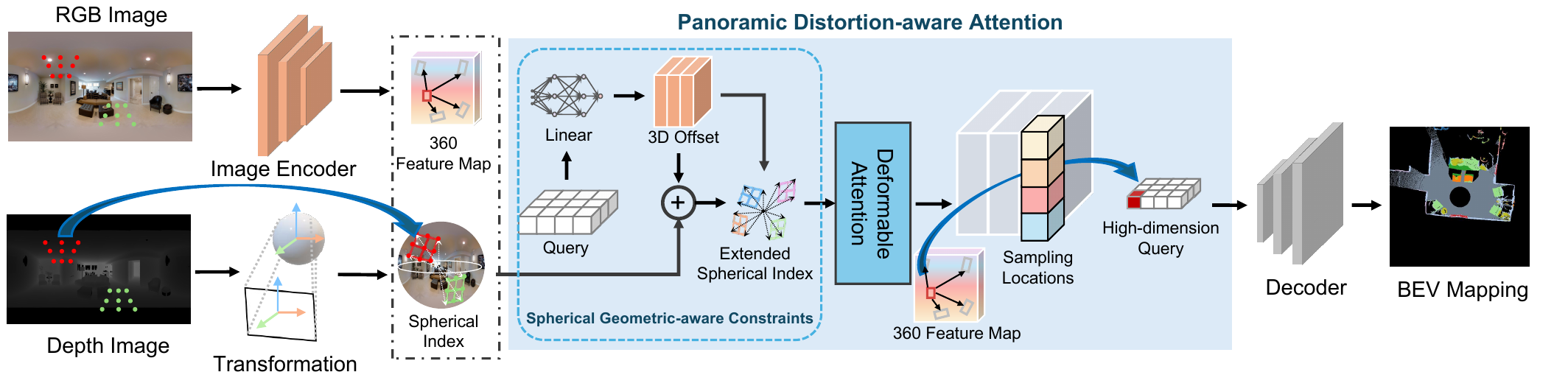} \\
		
	\end{tabular}
	\caption{\textbf{HumanoidPano Framework Overview.} The system addresses panoramic image distortion using Spherical Geometric Constraints (SGC) to guide 3D adaptive sampling offsets. By encoding camera ray properties into offset prediction, SGC aligns panoramic features with depth measurements, reducing projection artifacts. Spatial Deformable Attention (SDA) adaptively aggregates geometrically consistent object representations from panoramic images.
 }
	\label{fig:hpano_framework}
	\vspace{-0.35cm}
\end{figure*}
\section{Related Work}

\subsection{Humanoid Robot Perception}
In recent years, humanoid robots have garnered significant attention due to their human-like design and remarkable versatility in both living and working environments. To enhance the adaptability of humanoid locomotion and manipulation, it is essential to equip these robots with advanced perceptive abilities.

In the field of locomotion, existing methods primarily rely on elevation maps~\cite{miki2022elevation} and depth images. Studies such as \cite{zhuang2024humanoid, wang2025beamdojo, long2024learning,ren2025vb} employ LiDAR or depth cameras to construct elevation maps, providing sampled height points as observations for the humanoid. Similarly, \cite{duan2024learning} directly utilize egocentric depth images as perceptive input and leverage sampled height points from a simulator to supervise a height map predictor, facilitating improved sim-to-real transfer. However, these perception modalities primarily serve locomotion policies and thus contain only geometric information, lacking semantic understanding.

For manipulation, perception modalities are typically more diverse and offer greater accuracy. Approaches such as \cite{fu2024humanplus, he2024omnih2o, he2024learning, kim2024armor} use RGB cameras and teleportation devices to collect human demonstration data, enabling humanoid robots to autonomously complete tasks via imitation learning. \cite{li2024okami} further advances humanoid manipulation by leveraging RGB-D video demonstrations for skill acquisition. Additionally, \cite{ze2024generalizable} integrate a 3D LiDAR sensor with an enhanced diffusion policy imitation learning framework, allowing humanoid robots to perform tasks autonomously in diverse real-world scenarios.~\cite{zhu2025vr} build a navigation simulation environment with egocentric view RGB camera.

Despite these advancements, existing methods are constrained by their limited field of view and lack the capability to understand and interact with large-scale surroundings. To address this limitation, we propose HumanoidPano, a novel framework that, for the first time, fuses panoramic cameras and LiDAR to achieve comprehensive environmental awareness.

\subsection{Panoramic Perception}
With the rapid advancement of robotics and artificial intelligence, human perception of real-world scenes is no longer restricted to narrow fields of view and low-dimensional sensing devices. Panoramic perception enables more comprehensive environmental understanding and measurement. HoHoNet~\cite{sun2021hohonet} introduces a framework for indoor 360-degree panorama understanding using a Latent Horizontal Feature, achieving dense depth estimation, layout reconstruction, and semantic segmentation. \cite{wang2025depth} enhance monocular depth estimation for 360-degree images, enabling effective knowledge transfer across different camera projections and data modalities. Several works\cite{yun2023egformer,shen2022panoformer,li2022omnifusion} incorporate global attention mechanisms, significantly improving 360-degree depth estimation.

Semantic segmentation is also crucial for navigation and scene understanding. Research efforts such as~\cite{zheng2024360sfuda++,zhang2022bending,zheng2024semantics,dong2024panocontext,ling2023panoswin,yu2023panelnet,jiang2022lgt} provide a more comprehensive interpretation of indoor scenes through panoramic semantic segmentation. Additionally,~\cite{teng2024360bev} extend panoramic perception to generate BEV (Bird’s Eye View) segmentation maps. However, despite these advancements, many of these methods are constrained by the lack of real-world validation, as they primarily rely on limited datasets~\cite{chang2017matterport3d,zioulis2019spherical,albanis2021pano3d}.

\section{HumanoidPano}

Motivated by the critical need for embodied scene understanding in humanoid robotics, we formalize the perceptual challenge as Panoramic-LiDAR Cross-modal BEV Semantic Segmentation, a novel task requiring joint interpretation of 360° visual context and 3D spatial awareness under dynamic morphological constraints. Subsequently our proposed method will be shown in detail.

\subsection{Problem Formulation}
In this work, we aim to obtain accurate semantic BEV map from panoramic image for humanoid robots. Formally, it can be defined as:
\begin{equation}
\label{eq1}
M = F_{seg}(I, D)
\end{equation}

Where $F_{seg}$ is the segmentation model, $I \in R^{3 \times h \times w}$ is the input panoramic image, $D$ represent the depth image, $(h, w)$ represent the height and width of the image, $M \in R^{L \times X \times Y}$ is the output semantic BEV map, $(X, Y)$ represent the perception range of the BEV space, $L$ is the number of categories predicted.

\subsection{Overall Architecture}
The proposed HumanoidPano framework as shown in Figure \ref{fig:hpano_framework} includes three key components: (1) The transformer-based panoramic image encoder and transformation module. (2) The Panoramic Distortion-aware Attention (PD Attention) module which accomplishes the view transformation from panoramic image to BEV space and adaptive spatial feature extraction with spherical geometry constraints. (3) The BEV decoder parses the projected feature map and predicts the semantic BEV map.
\subsubsection{Panoramic Distortion-aware Attention}
Due to the huge and distorted panoramic image observed by the humanoid robot, it is difficult for the original 360 attention to learn the real 3D spatial distribution and complete representation information of objects with varying scales and distortions in panoramic images. Therefore, PD Attention is proposed, which learns feature aberrations in panoramic images through Spherical Geometry-aware Constraints (SGC) and grasps complete object information in real 3D space through Spatial Deformable Attention (SDA).\\
\textbf{Spherical Geometry-aware Constraints.} Given the BEV queries $Q \in R^{N\times C}$ and the 3D reference points $P \in R^{N\times 3}$ as input, where $N$ is the length of BEV queries and 3D reference points. The 3D reference points $P$ are obtained by projecting the depth map $D$ into the pseudo-point cloud. The $mask(·)$ operation is applied on $Q$ and $P$ to mask out irrelevant points. The 3D reference points $P$ are transformed onto the sphere to generate spherical reference points $S \in R^{N\times 2}$, as shown in Eq. (\ref{eq2}):
\begin{equation}
\label{eq2}
\Phi = tan^{-1}\frac{y}{x}, \Theta = tan^{-1}(\frac{x}{z}\cdot\frac{1}{cos(\Phi)})
\end{equation}

Where $(x, y, z)$ represent 3D reference coordinates in cartesian coordinate system, $(\Phi, \Theta)$ represent azimuthal angle and polar angle in spherical coordinate system. The spherical sampling offsets $\Delta S \in R^{N\times 2}$ are obtained by BEV queries with linear layer, which generates spherical locations $S' \in R^{N\times 2}$ with the original spherical reference points $S \in R^{N\times 2}$, as shown in Eq. (\ref{eq3}, \ref{eq4}): 
\begin{equation}
\label{eq3}
(\Delta\Phi, \Delta\Theta) = f_{\Delta s}(mask(Q) + f_{pos}(mask(S)))
\end{equation}
\begin{equation}
\label{eq4}
\Phi' = \Phi + \Delta\Phi, \Theta' = \Theta + \Delta\Theta  
\end{equation}

Where $f_{\Delta s}$ represent the linear layer for spherical angular deviations prediction and 
$f_{pos}$ represent the linear layer for spherical position encoding. Afterwards, we generate the 2D index $I \in R^{N\times 2}$ of the panoramic image from spherical locations $S'$, as shown in Eq. (\ref{eq5}):
\begin{equation}
\label{eq5}
I_h = \lceil \frac{H\Theta'}{\pi} \rceil, I_w = \lceil (\frac{\Phi'}{\pi} - \frac{1}{W})\cdot\frac{W}{2}\rceil
\end{equation}

Where $(H, W)$ represent height and width of panoramic image feature map, $(I_h, I_w)$  represent the 2D index values of spherical locations $S'$ on the panoramic image feature map.\\
\textbf{Spatial Deformable Attention.} We regard these 2D indexes $I$ as the reference points of the BEV queries $Q$ and sample the features from the panoramic image feature map $F \in R^{C\times H \times W}$. Finally, we perform a weighted sum of the sampled features as the output of SDA. The process of SDA can be formulated as:
\begin{equation}
\label{eq6}
SDA(Q, I, F) = \sum_{i}^{N_{ref}}A_{i}\cdot F[I+f_{\Delta ref}(mask(Q))]
\end{equation}

Where $f_{\Delta ref}$ represent the linear layer for 2D offsets prediction and $A_{i}$ represent the attention weight of each sampled point.

\subsubsection{Panoramic-LiDAR Cross-Modal Perception}

In real-world robotic scenarios, depth information is typically acquired using two primary sensor types: 3D structured light depth cameras and LiDAR. We adopt panoramic LiDAR cross modal perception due to limitations of conventional structured light depth cameras in real-world robotic applications. These sensors often fail to provide the large range detection required for humanoid navigation, whereas LiDAR offers extended sensing capabilities. Although LiDAR generates sparser depth maps compared to structured light, its depth information can be leveraged for sparse semantic queries, which is critical for scene understanding in occluded environments.

\begin{figure*}[t]
    \centering
    \includegraphics[width=2.1\columnwidth]{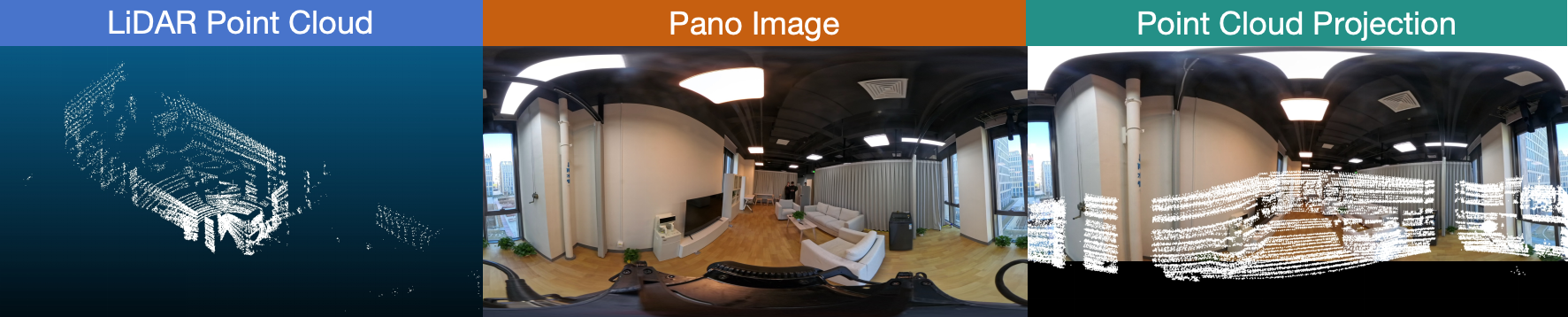}
    \caption{\textbf{Visualization of LiDAR Depth Extraction and Projection Process.} First, LiDAR captures raw point clouds, which are then projected onto panoramic images using the camera’s intrinsic and extrinsic parameters. This projected depth information is fed into the network, enabling the retrieval of pixel-wise semantics corresponding to the depth values.}
    \label{fig:lidarvis}
    \vspace{-0.35cm}
\end{figure*}

The primary distinction between LiDAR depth and structured light depth lies in data density. However, this sparsity does not compromise the framework's applicability. Our architecture is designed to inherently accommodate this characteristic through geometric constraint mechanisms, enabling seamless integration of panoramic and LiDAR modalities. This approach ensures robust environmental perception while maintaining computational efficiency for real-time humanoid operations.

\begin{figure}[t]
    \centering
    \includegraphics[width=\columnwidth]{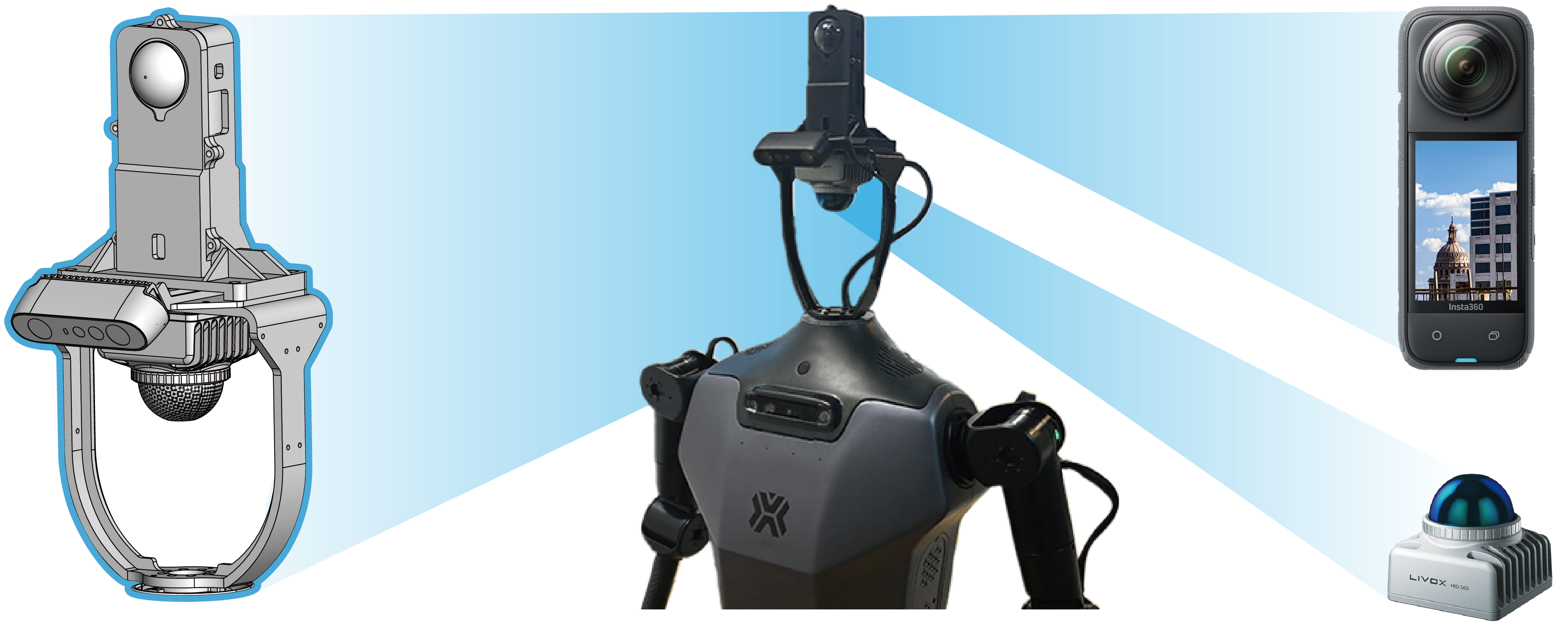}
    \caption{\textbf{Hardware design of our universal sensor module}, integrating an Insta360 X4 panoramic camera and Livox Mid-360 LiDAR with precise spatial alignment. The compact, lightweight assembly minimizes self-occlusion while maximizing 360° coverage. Its modular design supports flexible deployment across humanoid platforms for real-time panoramic-LiDAR fusion in navigation and manipulation tasks.}
    \label{fig:hardware}
    \vspace{-0.35cm}
\end{figure}

The paradigm of panoramic LiDAR cross modal perception is validated using real-world physical data. Specifically, the Insta360 X4 panoramic camera and Livox Mid360 LiDAR are utilized for data collection, with precise alignment achieved via timestamp synchronization. The point cloud data is transformed into the panoramic camera's coordinate system using an extrinsic calibration matrix, facilitating the construction of BEV-based map mask and map heights. HumanoidPano employs the PD Attention mechanism to generate 3D spatial points and 2D index values, subsequently performing inference to produce the final BEV semantic segmentation map. This study not only substantiates the effectiveness and robustness of the method in real-world scenarios but also pioneers new directions for panoramic vision and LiDAR fusion perception, expanding its applicability to autonomous navigation and beyond. Building upon our discussion of humanoid robot perception systems above, we present our hardware design in Figure \ref{fig:hardware}, which features a universal multimodal perception module. The projection visualization process of LiDAR point clouds and panoramic images is presented in Figure \ref{fig:lidarvis}.

\subsection{Data Augmentation}
The benefits of data augmentation have formed a consensus in 2D image processing.We designed a data augmentation method for joint panoramic image space and BEV space, which is still not effectively explored. Visualizations and more details of data augmentation are presented in the appendix materials.\\
\textbf{Panoramic Image Augmentation.}
For a panoramic image, to enhance its diversity, we randomly flip it and randomly mix it with another panoramic image in the training set.\\
\textbf{BEV Augmentation.}
Similar to panoramic image augmentation, BEV augmentation imposes the random flipping and mixing described above. Note that data augmentation in the BEV space is more challenging than the image space because the image space and BEV space need to be aligned for proper view transformation. Therefore, we need to apply corresponding rotation and mixing to the ground truth of the semantic map in order to align the panoramic image augmentation.

\section{Experiments}

In this section, we comprehensively present the performance of our proposed approach across multiple datasets. We begin by introducing the prevalent panoramic camera perception datasets in Section \ref{dataset}, which serve as the basis for training and  evaluating our method. Subsequently, in Section \ref{semantic resluts}, we demonstrate the high pixel wise accuracy, recall, and mean Intersection-over-Union (mIoU) achieved by our method on these datasets, highlighting its effectiveness in semantic segmentation tasks. Additionally, we conduct an ablation study in Section \ref{ablation} to analyze the contributions of different components of our method.

Regarding the real world deployment on a humanoid robot, considering its unique requirements, we have designed a specialized layout and sensor arrangement, details of which are presented in Section \ref{real-robot}. Given the limited sensing range and field of view (FOV) of structured - light depth cameras in real world scenarios, we opted to use a LiDAR for depth information acquisition. Finally, validation demos carried out on an actual humanoid robot platform convincingly demonstrate the overall effectiveness of our solution, highlighting its potential for practical applications in humanoid robotics.

\subsection{Dataset}
\label{dataset}

\begin{figure}[t]
    \centering
    \includegraphics[width=\columnwidth]{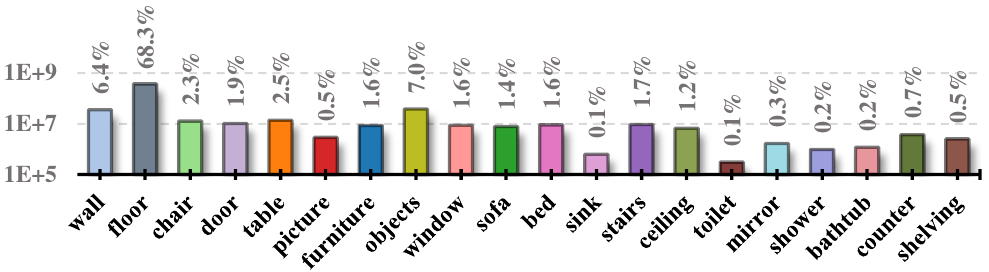}
    \caption{\textbf{The data distribution of 360BEV-Matterport visualized by category\cite{teng2024360bev}.} We directly adopted the well-established indoor classification scheme and visual schematics from 360BEV-Matterport for dataset presentation, these categories enable humanoid robots to effectively perform indoor navigation tasks.}
    \label{fig:dataset}
    \vspace{-0.35cm}
\end{figure}

The scarcity of panoramic datasets, compounded by the challenges of spherical image annotation, poses significant limitations to model generalization in panoramic perception research. However, we address this gap by leveraging the 360BEV-Matterport dataset\cite{teng2024360bev}, a tailored benchmark uniquely aligned with humanoid robotic perception requirements. Derived from Matterport3D, this dataset provides panoramic RGB images (512×1024 resolution) paired with dense BEV semantic annotations for 20 navigation-critical categories (Figure \ref{fig:dataset}). Synthetic depth maps generated via ray casting ensure geometric consistency with panoramic inputs, enabling rigorous validation of our cross-modal fusion framework.
The dataset's multi-scenario coverage of cluttered indoor environments (living spaces, offices) closely mirrors real-world humanoid operation scenarios. While BEV annotations directly align with our goal of generating navigable semantic maps. The spherical projection of panoramic images matches humanoid sensor configurations, and the combination of 360° visual context, synthetic depth, and navigation-centric labels makes this dataset ideal for evaluating our distortion-aware panoramic-LiDAR fusion approach.


\subsection{Training Details}


We adopted the training settings of 360BEV\cite{teng2024360bev}. Specifically, we trained the HumanoidPano models using 4 A100 GPUs. The initial learning rate was set to $6\times10^{-5}$, and the step strategy was employed for scheduling over 50 epochs. We utilized the AdamW optimizer with an epsilon value of $1\times10^{-8}$, a weight decay of 0.01, and a batch size of 4 on each GPU.

The panoramic images in 360FV-Matterport\cite{teng2024360bev} dataset had a size of $512\times1024$. For the HumanoidPano training, the panoramic images from the 360BEV-Stanford and 360BEV-Matterport datasets, which also had a resolution of $512\times1024$, were used as inputs. Meanwhile, the output BEV maps were configured to be $500\times500$, corresponding to a perception range of $10\text{m}\times10\text{m}$.
Following the works of \cite{cartillier2021semantic, chen2023trans4map}, we employed pixel-wise accuracy (Acc), pixel recall (mRecall), precision (mPrecision), and mean Intersection-over-Union (mIoU) as the evaluation metrics.

\begin{table}[h]
\large
\renewcommand{\arraystretch}{1.2}
\setlength{\heavyrulewidth}{1.2pt}
  \caption{Panoramic semantic mapping on the val set of 360 BEV-Matterport dataset. We marked the data of HumanoidPano with \textcolor{lightblue}{light blue}. Meanwhile, we marked the state-of-the-art (SOTA) results with \textcolor{iccvblue}{dark blue} and indicated the sub-optimal results with \underline{underlines}(on each backbone). \dag\  benefits from data augmentation.}
  \centering
  \label{table1}
  \begin{adjustbox}{width=\linewidth}
    \begin{tabular}{lcccccc}   
    \toprule
    Method & Proj & Backbone & Acc & mRecall & mPrecision & mIoU\\
    \midrule
    SegFormer \cite{yun2023egformer}&E&MiT-B2&68.12&41.33&45.25&29.22\\
    SegNeXt \cite{guo2022segnext}&E&MSCA-B&68.53&42.13&46.12&30.01\\
    \midrule
    HoHoNet \cite{sun2021hohonet}&L&ResNet101&62.84&38.99&44.22&26.21\\
    Trans4PASS \cite{zhang2022bending}&L&MiT-B2&55.99&29.59&40.91&20.07\\
    Trans4PASS+ \cite{zhang2024behind}&L&MiT-B2&57.89&32.75&40.93&21.58\\
    SegFormer \cite{yun2023egformer}&L&MiT-B2&62.98&41.84&45.30&27.78\\
    \midrule
    Trans4Map \cite{chen2023trans4map}&I&MiT-B0&70.19&44.31&50.39&31.92\\
    360BEV \cite{teng2024360bev}&I&MiT-B0&74.72&47.14&55.20&35.46\\
    \rowcolor{lightblue}
    HumanoidPano&I&MiT-B0& \underline{76.19}& \underline{50.41}&\textbf{\textcolor{iccvblue}{57.86}}& \underline{38.50}\\
    \rowcolor{lightblue}
    HumanoidPano \dag&I&MiT-B0&\textbf{\textcolor{iccvblue}{76.71}}&\textbf{\textcolor{iccvblue}{52.87}}& \underline{57.76}&\textbf{\textcolor{iccvblue}{40.12}}\\
    BEVFormer \cite{li2024bevformer}&I&MiT-B2&72.99&43.61&51.70&32.51\\
    Trans4Map \cite{chen2023trans4map}&I&MiT-B2&73.28&51.60&53.02&36.72\\
    360BEV \cite{teng2024360bev}&I&MiT-B2&76.98&54.30&59.63&41.42\\
    \rowcolor{lightblue}
    HumanoidPano&I&MiT-B2& \underline{77.72}& \underline{57.54}& \underline{60.18}& \underline{43.52}\\
    \rowcolor{lightblue}
    HumanoidPano \dag&I&MiT-B2&\textbf{\textcolor{iccvblue}{78.48}}&\textbf{\textcolor{iccvblue}{58.32}}&\textbf{\textcolor{iccvblue}{61.52}}&\textbf{\textcolor{iccvblue}{44.54}}\\
    360BEV \cite{teng2024360bev}&I&MSCA-B&78.37&58.08&62.62&45.04\\
    \rowcolor{lightblue}
    HumanoidPano&I&MSCA-B&\underline{79.21}&\underline{60.41}&\textbf{\textcolor{iccvblue}{63.17}}&\textbf{\textcolor{iccvblue}{46.47}}\\
    \rowcolor{lightblue}
    HumanoidPano \dag&I&MSCA-B&\textbf{\textcolor{iccvblue}{79.47}}&\textbf{\textcolor{iccvblue}{60.92}}&\underline{62.66}&\underline{46.35}\\
    \bottomrule 
    \end{tabular}
 \end{adjustbox}
\end{table}

\begin{table}[h]
\large
\renewcommand{\arraystretch}{1.2}
\setlength{\heavyrulewidth}{1.2pt}
  \caption{Panoramic semantic mapping on the test set of 360 BEV-Matterport dataset. The annotation is consistent with Table \ref{table1}.}
  \centering
  \label{table2}
  \begin{adjustbox}{width=\linewidth}
    \begin{tabular}{lcccccc}
    \toprule
    Method & Proj & Backbone & Acc & mRecall & mPrecision & mIoU\\
    \midrule
    SegFormer \cite{yun2023egformer}&E&MiT-B2&69.72&35.28&40.41&24.04\\
    SegNeXt \cite{guo2022segnext}&E&MSCA-B&69.99&36.25&41.96&25.22\\
    \midrule
    HoHoNet \cite{sun2021hohonet}&L&ResNet101&62.89&35.18&39.54&22.01\\
    Trans4PASS \cite{zhang2022bending}&L&MiT-B2&53.50&29.35&33.53&16.53\\
    Trans4PASS+ \cite{zhang2024behind}&L&MiT-B2&57.24&30.64&34.49&17.72\\
    SegFormer \cite{yun2023egformer}&L&MiT-B2&62.91&35.35&39.64&22.02\\
    \midrule
    Trans4Map \cite{chen2023trans4map}&I&MiT-B0&71.78&38.27&43.77&26.52\\
    360BEV \cite{teng2024360bev}&I&MiT-B0&75.98&44.75&48.59&31.81\\
    \rowcolor{lightblue}
    HumanoidPano&I&MiT-B0& \underline{76.48}& \underline{47.09}& \underline{51.42}& \underline{33.56}\\
    \rowcolor{lightblue}
    HumanoidPano \dag&I&MiT-B0&\textbf{\textcolor{iccvblue}{77.50}}&\textbf{\textcolor{iccvblue}{49.26}}&\textbf{\textcolor{iccvblue}{52.23}}&\textbf{\textcolor{iccvblue}{35.37}}\\ 
    BEVFormer \cite{li2024bevformer}&I&MiT-B2&72.04&36.69&47.90&27.46\\
    Trans4Map \cite{chen2023trans4map}&I&MiT-B2&72.94&45.45&47.03&31.08\\
    360BEV \cite{teng2024360bev}&I&MiT-B2&77.95&50.55&53.06&36.00\\
    \rowcolor{lightblue}
    HumanoidPano&I&MiT-B2& \underline{78.02}& \underline{53.22}& \underline{53.93}& \underline{37.95}\\
    \rowcolor{lightblue}
    HumanoidPano \dag&I&MiT-B2&\textbf{\textcolor{iccvblue}{78.16}}&\textbf{\textcolor{iccvblue}{53.60}}&\textbf{\textcolor{iccvblue}{54.81}}&\textbf{\textcolor{iccvblue}{38.49}}\\
    360BEV \cite{teng2024360bev}&I&MSCA-B&\textbf{\textcolor{iccvblue}{79.07}}&53.59&56.40&39.10\\
    \rowcolor{lightblue}
    HumanoidPano&I&MSCA-B&\underline{78.96}&\underline{55.64}&\underline{56.61}&\underline{40.27}\\
    \rowcolor{lightblue}
    HumanoidPano \dag&I&MSCA-B&78.92&\textbf{\textcolor{iccvblue}{55.69}}&\textbf{\textcolor{iccvblue}{56.86}}&\textbf{\textcolor{iccvblue}{40.41}}\\
    \bottomrule 
    \end{tabular}
 \end{adjustbox}
\end{table}

\begin{table}[t]
\large
\renewcommand{\arraystretch}{1.2}
\setlength{\heavyrulewidth}{1.2pt}
  \caption{Results under noise with backbone MiT-B0 on the val set of 360 BEV-Matterport dataset. The annotation is consistent with Table \ref{table1}.}
  \centering
  \label{table3}
  \begin{adjustbox}{width=\linewidth}
    \begin{tabular}{lcccccc}
    \toprule
    Method&Pitch&Roll& Acc & mRecall & mPrecision & mIoU\\
    \midrule
    360BEV&-&-&74.72&47.14&55.20&35.46\\
    360BEV&\checkmark&-&69.85&40.87&49.11&29.74\\
    360BEV&-&\checkmark&69.74&40.75&49.19&29.64\\
    360BEV&\checkmark&\checkmark&66.65&37.82&44.97&26.67 (-24.78$\%$)\\
    \midrule
    HumanoidPano&-&-&76.19&50.41&57.86&38.50\\
    HumanoidPano&\checkmark&-&71.46&44.01&51.85&32.84\\
    HumanoidPano&-&\checkmark&71.50&43.81&51.97&32.79\\
    \rowcolor{lightblue}
    HumanoidPano&\checkmark&\checkmark&68.41&40.32&47.87&29.58 (-\underline{23.16}$\%$)\\
    \midrule
    HumanoidPano \dag&-&-&76.71&52.87&57.76&40.12\\
    HumanoidPano \dag&\checkmark&-&72.12&46.67&51.50&34.28\\
    HumanoidPano \dag&-&\checkmark&72.35&46.64&52.21&34.55\\
    \rowcolor{lightblue}
    HumanoidPano \dag&\checkmark&\checkmark&69.19&43.25&48.27&31.13 (-\textbf{\textcolor{iccvblue}{22.40}}$\%$)\\
    \bottomrule
    \end{tabular}
    \end{adjustbox}
\end{table}

\begin{table}[t]
\large
\renewcommand{\arraystretch}{1.2}
\setlength{\heavyrulewidth}{1.2pt}
  \caption{Ablation study on the test set of 360 BEV-Matterport dataset. The annotation is consistent with Table \ref{table1}.}
  \centering
  \label{table4}
  \begin{adjustbox}{width=\linewidth}
    \begin{tabular}{cccccccc}
    \toprule
    SDA&SGC&AUG&Backbone & Acc & mRecall & mPrecision & mIoU\\
    \midrule
    -&-&-&MiT-B0&75.98&44.75&48.59&31.81\\
    \checkmark&-&-&MiT-B0&76.38&46.78&49.72&33.05\\
    \checkmark&\checkmark&-&MiT-B0&76.48&47.09&51.42&33.56\\
    \rowcolor{lightblue}
    \checkmark&\checkmark&\checkmark&MiT-B0&\textbf{\textcolor{iccvblue}{77.50}}&\textbf{\textcolor{iccvblue}{49.26}}&\textbf{\textcolor{iccvblue}{52.23}}&\textbf{\textcolor{iccvblue}{35.37}}\\
    \bottomrule 
    \end{tabular}
    \end{adjustbox}
\end{table}

\begin{table*}[t]
\large
\renewcommand{\arraystretch}{1.2}
\setlength{\heavyrulewidth}{1.2pt}
  \caption{Per-class results with backbone MiT-B2 on the 360BEV-Matterport dataset. The annotation is consistent with Table \ref{table1}.}
  \centering
  \label{table5}
  \begin{adjustbox}{width=\linewidth}
    \begin{tabular}{lccccccccccccccccccccccc}
    \toprule
    Method&Data&\rotatebox[origin=c]{45}{\textbf{mIoU}}&\rotatebox[origin=c]{45}{void}&\rotatebox[origin=c]{45}{wall}&\rotatebox[origin=c]{45}{floor}&\rotatebox[origin=c]{45}{chair}&\rotatebox[origin=c]{45}{door}&\rotatebox[origin=c]{45}{table}&\rotatebox[origin=c]{45}{picture}&\rotatebox[origin=c]{45}{furniture}&\rotatebox[origin=c]{45}{objects}&\rotatebox[origin=c]{45}{window}&\rotatebox[origin=c]{45}{sofa}&\rotatebox[origin=c]{45}{bed}&\rotatebox[origin=c]{45}{sink}&\rotatebox[origin=c]{45}{stairs}&\rotatebox[origin=c]{45}{ceiling}&\rotatebox[origin=c]{45}{toilet}&\rotatebox[origin=c]{45}{mirror}&\rotatebox[origin=c]{45}{shower}&\rotatebox[origin=c]{45}{bathtub}&\rotatebox[origin=c]{45}{counter}&\rotatebox[origin=c]{45}{shelving}\\
    \midrule
    Trans4Map \cite{chen2023trans4map}&val&36.72&47.87&28.52&82.96&34.44&22.27&39.58&16.28&22.75&26.29&25.08&42.81&62.25&13.95&41.51&37.79&45.82&19.56&48.05&47.71&38.25&27.31\\
    360BEV \cite{teng2024360bev}&val&41.42&78.73&32.10&84.23&40.60&23.92&44.89   &18.17&22.30&22.98&25.17&48.30&64.94&16.95&44.01&49.48&54.51&19.12&50.71&59.01&43.93&25.85\\
    \rowcolor{lightblue}
    HumanoidPano&val&\underline{43.52}&73.34&33.72&85.94&41.74&25.00&47.07&20.98&26.20&23.59&27.63&49.35&66.52&21.50&45.29&52.80&57.12&27.53&49.87&60.93&48.90&28.93\\
    \rowcolor{lightblue}
    HumanoidPano \dag&val&\textbf{\textcolor{iccvblue}{44.54}}&76.36&33.29&85.95&42.98&27.34&47.18&21.74&23.65&28.85&27.56&50.81&65.09&19.60&48.11&58.15&54.49&26.90&51.53&66.79&45.58&33.28\\

    \midrule
    Trans4Map \cite{chen2023trans4map}&test&31.08&40.51&32.54&80.21&33.23&20.85&37.21&19.01&18.46&23.05&23.56&32.35&52.08&15.34&29.02&18.27&41.90&15.39&25.58&48.19&30.38&15.52\\
    360BEV \cite{teng2024360bev}&test&36.00&63.47&35.50&84.95&39.02&17.85&42.12&19.96&22.07&23.09&26.60&36.18&58.48&22.80&37.08&47.09&38.76&17.03&21.38&52.83&35.26&14.66\\
    \rowcolor{lightblue}
    HumanoidPano&test&\underline{37.95}&60.43&37.04&84.77&39.89&21.64&44.25&22.16&22.11&24.43&26.82&35.76&60.42&27.77&36.23&48.27&54.62&15.73&27.22&53.71&35.40&18.27\\
    \rowcolor{lightblue}
    HumanoidPano \dag&test&\textbf{\textcolor{iccvblue}{38.49}}&61.38&37.72&84.39&39.41&22.91&43.60&20.94&21.97&24.75&27.13&37.44&61.05&24.64&36.93&50.05&56.09&19.41&29.64&52.66&36.40&19.78\\
    \bottomrule
    \end{tabular}
    \end{adjustbox}
\end{table*}

\subsection{Panorama Semantic Mapping}
\label{semantic resluts}
\begin{figure}[t]
    \centering
    \includegraphics[width=\columnwidth]{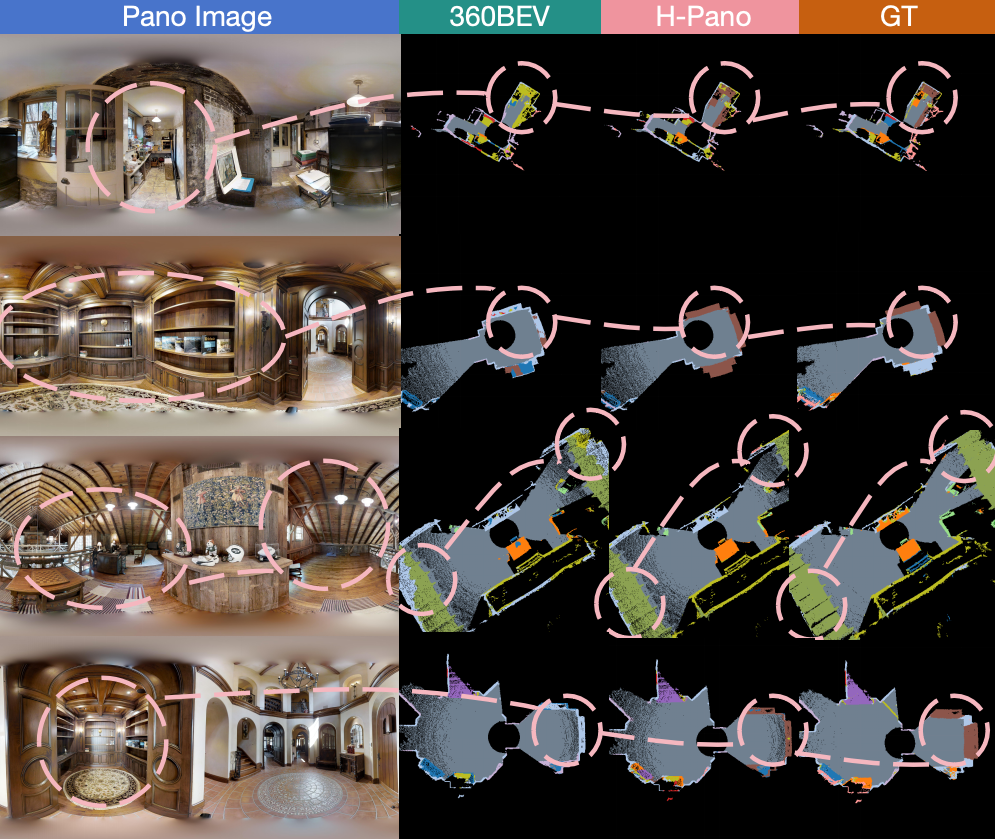}
    \caption{\textbf{The comparative experiments on the 360BEV-Matterport dataset.} We compared the inference results of 360BEV \cite{teng2024360bev} and our model, where HumanoidPano demonstrated superior performance in areas with distortions.}
    \label{fig:compare}
    \vspace{-0.35cm}
\end{figure}

We compare our approach with three backbones, e.g., MiT-B0, MiT-B2 from SegFormer \cite{yun2023egformer} and MSCA-B from SegNeXt \cite{guo2022segnext} for the panoramic semantic mapping task. Where the E represents the early projection methods, the L represents the late projection methods and the I represents the intermediate projection methods. Intermediate projection can preserve dense visual cues and long-range information from front-view panoramas, and deliver more valuable context for BEV semantic mapping, leading to this superiority, as compared to the other paradigms. As visualized in Figure \ref{fig:compare}, our method outperforms baselines in distorted panoramic regions—especially on severely warped objects—a capability vital for panoramic perception in humanoid and general embodied AI systems.\\
\textbf{Results on 360BEV-Matterport.}
We conduct extensive experiments on the val and test set of 360BEV-Matterport and compare our HumanoidPano with other state-of-the-art methods like BEVFormer \cite{li2024bevformer}, Trans4Map \cite{chen2023trans4map} and 360BEV \cite{teng2024360bev} to validate the effectiveness of the proposed HumanoidPano. In 360BEV-Matterport val set, as can be seen in Table \ref{table1}, the proposed HumanoidPano with MiT-B0 surpasses 360BEV by a significant margin of 1.99$\%$, 5.73$\%$, 2.56$\%$ and 4.66$\%$ in Acc, mRecall, mPrecision and mIoU, respectively. The proposed HumanoidPano with MiT-B2 surpasses 360BEV by a significant margin of 1.50$\%$, 4.02$\%$, 1.89$\%$ and 3.12$\%$ in Acc, mRecall, mPrecision and mIoU, respectively. The proposed HumanoidPano with MSCA-B surpasses 360BEV by 0.84$\%$, 2.33$\%$, 0.55$\%$ and 1.43$\%$ in Acc, mRecall, mPrecision and mIoU, respectively. In 360BEV-Matterport test set, as can be seen in Table \ref{table2}, the proposed HumanoidPano with MiT-B0 surpasses 360BEV by a significant margin of 1.52$\%$, 4.51$\%$, 3.64$\%$ and 3.56$\%$ in Acc, mRecall, mPrecision and mIoU, respectively. The proposed HumanoidPano with MiT-B2 surpasses 360BEV by a significant margin of 0.21$\%$, 3.05$\%$, 1.75$\%$ and 2.49$\%$ in Acc, mRecall, mPrecision and mIoU, respectively. The proposed HumanoidPano with MSCA-B surpasses 360BEV by 2.10$\%$, 0.46$\%$ and 1.31$\%$ in mRecall, mPrecision and mIoU, respectively.\\
\textbf{Results on noisy settings.}
In the real world, the panoramic image of humanoid robots are susceptible to change by disturbances on camera. Here, we evaluate the performance of the proposed method in such noisy settings. Specifically, we transform the panoramic image to the spherical coordinate system and then apply a rotation along the pitch and roll directions. The corresponding rotation matrix is $R(\alpha, \beta, 0)$, where the $\alpha$ is pitch angle and the $\beta$ is roll angle. As shown in Table \ref{table3}, the performance of the existing methods degrades significantly when the panoramic image is disturbed. In such noise settings, the proposed HumanoidPano has less performance degradation and remains ahead compared to the 360BEV \cite{teng2024360bev} due to its ability to understand the 3D world. And the proposed HumanoidPano shows more robust performance in the data augmentation setting. Visulization results are presented in the appendix materials.\\
\textbf{Ablation Study.}
\label{ablation}
Experiments in Table \ref{table4} show the improvement when our Spatial Deformable Attention (SDA), Spherical Geometry-aware Constraints (SGC) and data augmentation (AUG) are applied. The proposed SDA complements the spatial context information and enhances the network's ability to understand the 3D world, obtaining a relatively significant improvement. The proposed SGC builds the panoramic perception capability and learns distortion features, forcing the SDA to better adapt to the structure of distorted objects in panoramic images. The proposed AUG creates several new scenes by random rotation and mixing, which enhances the diversity of the data, improves the robustness of the model, and also obtains a relatively significant improvement.\\
\textbf{Per-class Results.}
Per-class IoU scores on 360BEV-Matterport are shown in Table \ref{table5}. Compared to Trans4Map \cite{chen2023trans4map} and 360BEV \cite{teng2024360bev}, our HumanoidPano with the same MiT-B2 backbone can achieve respective 44.54$\%$ and 38.49$\%$ in mIoU on the val and test set. The void class is also included on the 360BEV-Matterport dataset.

\subsection{Implementation on Humanoid Robot}
\label{real-robot}

We deploy and validate the HumanoidPano system on Tiangong, a full-sized humanoid platform (1.65m height, 23 DoF), through comprehensive indoor environmental testing. Developed by the Beijing Humanoid Robot Innovation Center, Tiangong demonstrates advanced all-terrain mobility. Our decision to test on this highly capable all-terrain humanoid robot allows us to better demonstrate the performance of our perception system. The perception module integrates  an Insta360 X4 panoramic camera and Livox Mid-360
LiDAR, as illustrated in Figure \ref{fig:hardware}. This configuration achieves minimal structural interference through kinematic collision analysis, ensuring continuous 360° sensing coverage during dynamic motions such as crawling and stair navigation. A hardware-accelerated processing pipeline synchronizes spherical visual data with LiDAR point clouds at 10Hz by using a Nvidia Jetson Orin AGX, while an online calibration protocol dynamically adjusts extrinsic parameters to compensate for sensor displacement caused by joint movements.

Evaluations in cluttered domestic environments demonstrate the system's operational efficacy. HumanoidPano maintains robust BEV semantic mapping during complex full-body interactions, including doorway traversal and obstacle navigation. The framework successfully resolves system-level challenges in humanoid robots' omnidirectional semantic perception systems, addressing the long-standing issue of simultaneously acquiring depth information and semantic understanding. By leveraging efficient perception neural networks, it provides robust semantic guidance for downstream navigation modules as shown in Figure \ref{walk}. This implementation marks the first realization of embodied panoramic-BEV perception on a full-scale humanoid platform, proving its capability to support advanced navigation task in human-centric environments without external infrastructure dependencies. Video demonstration results are presented in the appendix materials.

\section{Conclusion}
This work pioneers large-scale semantic mapping for humanoid robots through the HumanoidPano framework, establishing the first perception system capable of real-time environmental comprehension across full-body motion scenarios. By addressing the fundamental conflict between biomimetic morphology and sensing requirements, we demonstrate that structural constraints can be transformed into perceptual advantages when algorithms are co-designed with robotic embodiment.
The methodology sets a new precedent for morphology-aware perception, providing both theoretical foundations and implementation guidelines for next-generation human-centered robotic systems.
In the future, we will further optimize the perception systems and algorithms for multi-view and panoramic cameras, while collecting and annotating scene data relevant to humanoid robots for further research.

\section{Acknowledgments}

The authors would like to express sincere gratitude to the engineers at Beijing Innovation Center of Humanoid Robot  for their invaluable technical expertise and dedicated collaboration throughout this research. Their profound insights were instrumental in tackling the challenges posed by the complex coupled tasks of the humanoid robot platform.

Special thanks are extended to Arashi Vision Inc. (Insta360) for their generous support and cutting-edge technological assistance, which significantly contributed to the experimental implementation and data acquisition processes.

This work would not have been possible without their professional commitment and interdisciplinary synergy.
{
    \small
    \bibliographystyle{ieeenat_fullname}
    \bibliography{main}
}

\clearpage
\appendix

\section{Appendix}
\label{appendix}
\label{appendix-data}
\begin{figure*}[t]
    \centering
    \includegraphics[width=2\columnwidth]{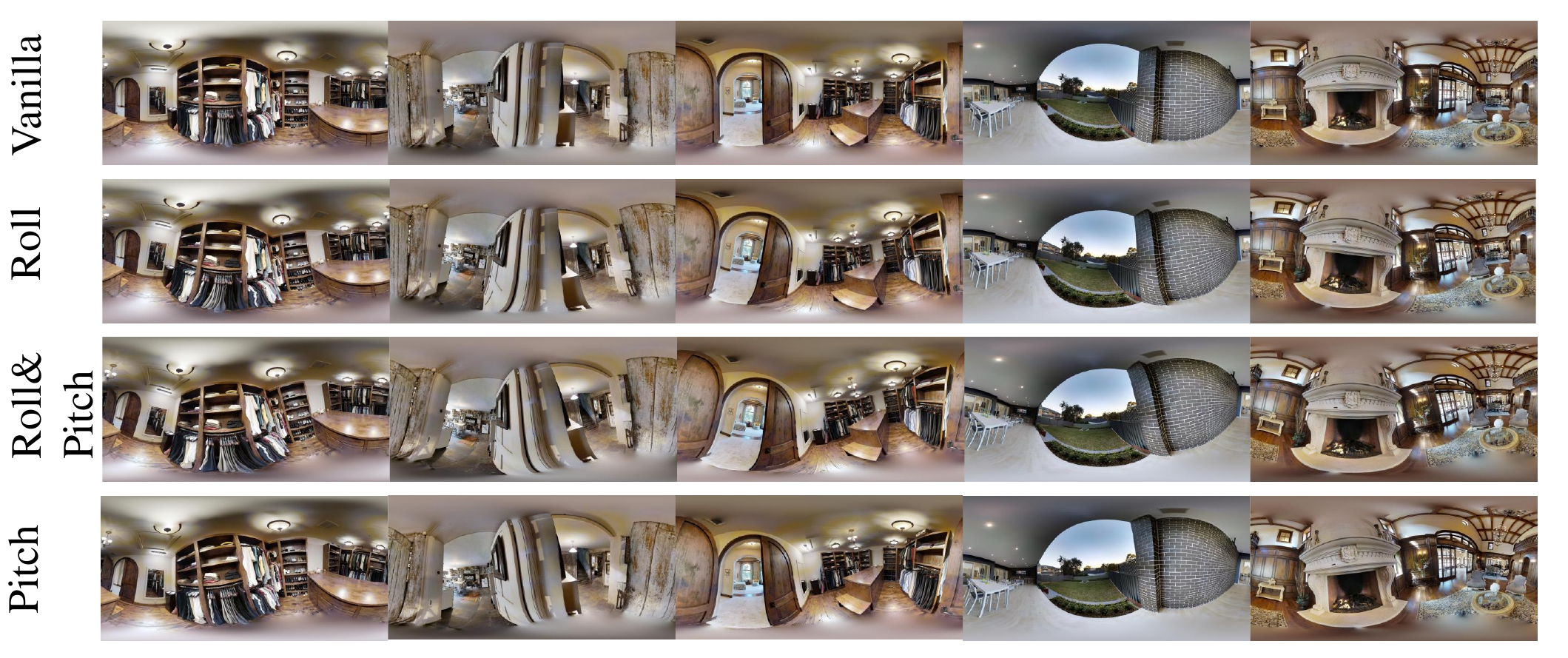}
    \caption{\textbf{Visualizations Panoramic Robustness Evaluation.} Roll and Pitch mean the rotation of images in the following direction.}
    \label{fig:rotation aug}
    \vspace{-0.35cm}
\end{figure*}

\begin{figure*}[t]
    \centering
    \includegraphics[width=2\columnwidth]{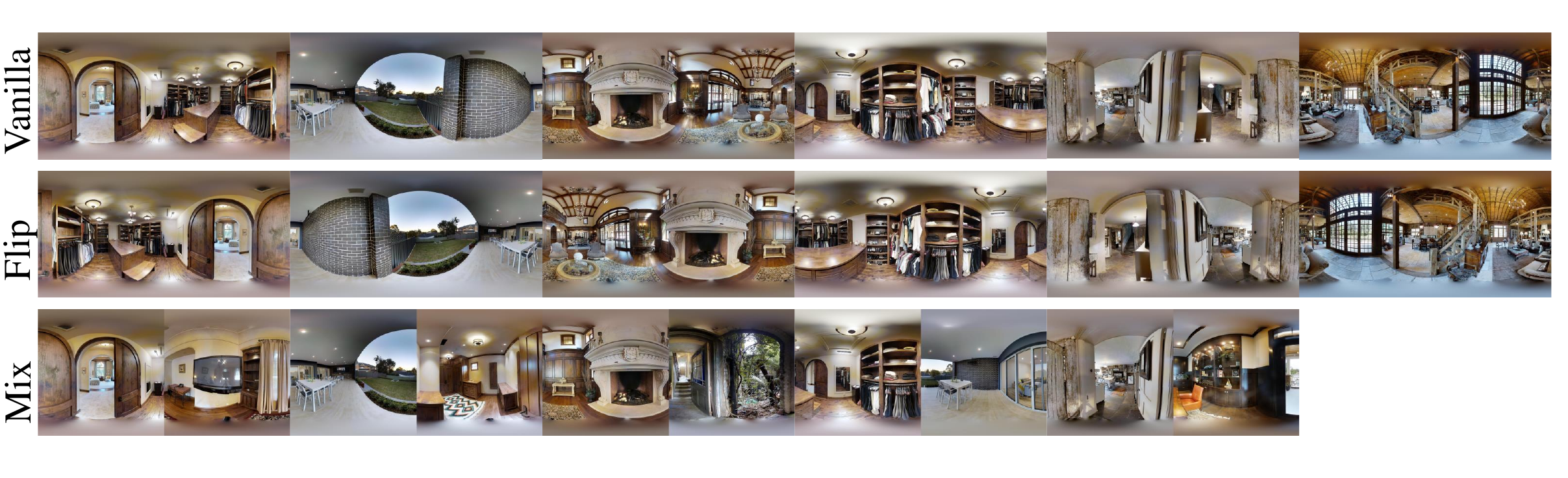}
    \caption{\textbf{Visualizations Panoramic Images Augmentation.} The horizontal flip and mix augmentation.}
    \label{fig:flip aug}
    \vspace{-0.35cm}
\end{figure*}

\begin{figure*}[t]
    \centering
    \includegraphics[width=1.5\columnwidth]{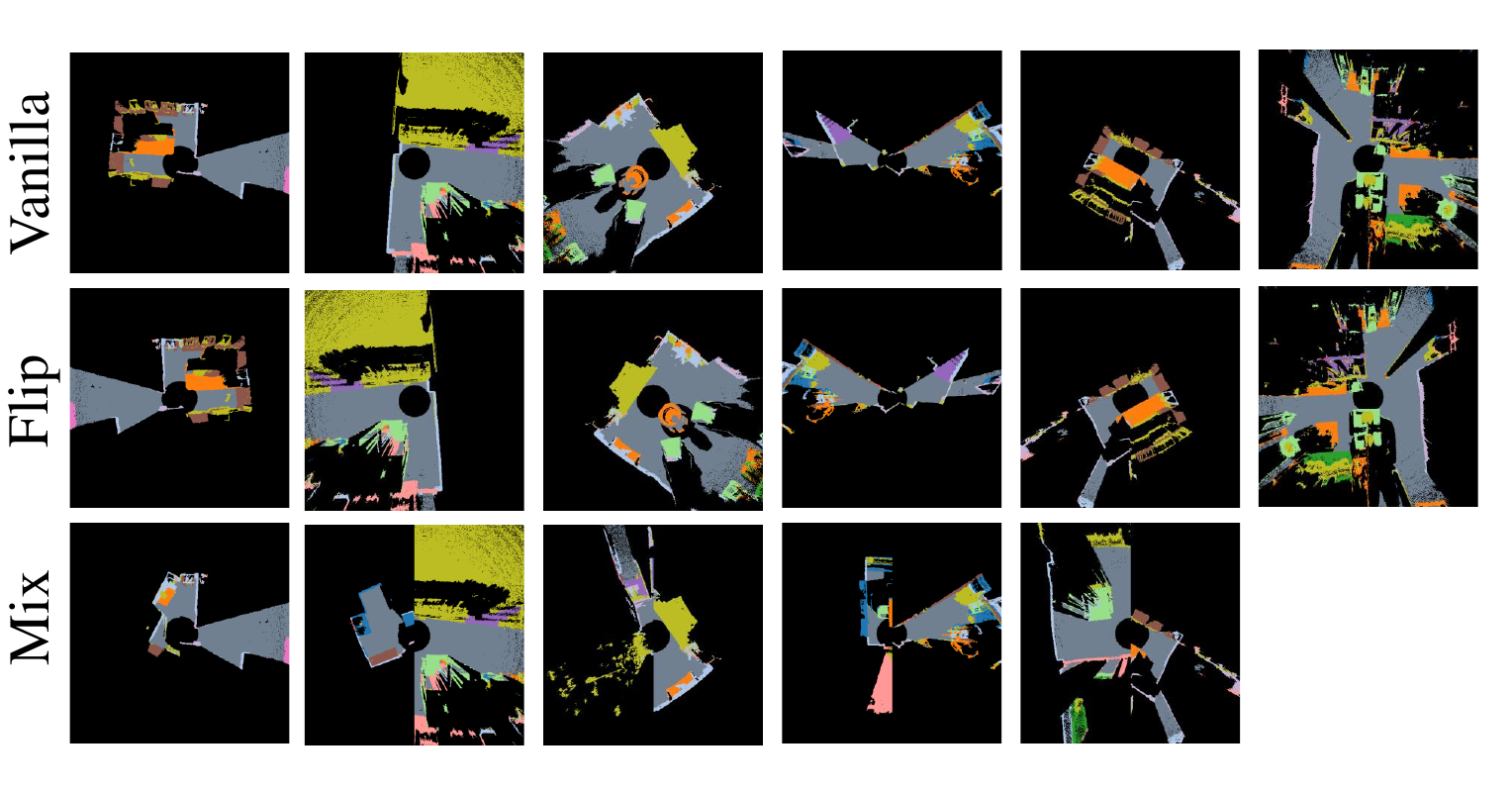}
    \caption{\textbf{Visualizations BEV Images Augmentation.} The horizontal flip and mix augmentation.}
    \label{fig:bevaug}
    \vspace{-0.35cm}
\end{figure*}

For evaluating the robustness of our methods, we rotate the panoramic image as showing in Figure~\ref{fig:rotation aug}.
The benefits of data augmentation have formed a consensus in 2D image processing. Augmenting data helps models generalize better by introducing variations in the training set, reducing overfitting, and improving robustness to unseen scenarios. The visualization of data augmentation for joint panoramic image space, as Figure~\ref{fig:flip aug}, Bird’s Eye View (BEV) space like Figure~\ref{fig:bevaug}.  The augmentation techniques incorporate spatial transformations and feature-mixing strategies to enhance generalization while maintaining geometric alignment. Visualizations of data augmentation are provided in the appendix materials.
For a panoramic image, data augmentation techniques are applied to increase diversity in training samples. Given a panoramic image 
$I$, a horizontal flip operation is applied with probability $p$. Let $W$ be the width of the image, and let $I(x,y)$ represent the pixel value at coordinate $(x,y)$. The flipped image $I'$ is defined as:
\begin{equation} I'(x, y) = I(W - x, y),  \quad \forall x \in [0, W], \quad \forall y \in [0, H], \end{equation}

where $H$ is the height of the image. The horizontal flip is applied as follows:

\begin{equation} I' =
\begin{cases} I(W - x, y), & \text{with probability } p, \\
I(x, y), & \text{otherwise}. \end{cases} \end{equation}

Since panoramic images wrap around at the boundaries, care must be taken to ensure continuity when flipping.

To introduce additional variations, a random mixing strategy is applied. Given two panoramic images $I_1$ and $I_2$ sampled from the training set, a mixing factor $\lambda$ is used to blend them as follows:

\begin{equation} I' = \lambda I_1 + (1 - \lambda) I_2. \end{equation}

This operation encourages the model to learn smooth transitions between different scenes and reduces sensitivity to specific environmental conditions.

BEV augmentation follows a similar approach to panoramic image augmentation, including random flipping and image mixing. However, data augmentation in the BEV space is more complex due to the need for geometric consistency between the panoramic and BEV representations. To ensure proper alignment, additional constraints are introduced. The flipping operation in the BEV space must correspond to the transformation applied in the panoramic space. Given a BEV semantic map 
$S$, the flipping function is defined as:

\begin{equation} S'(x, y) = S(W - x, y). \end{equation}

This ensures that any augmentation in the image space is accurately reflected in the BEV space.

A random rotation is applied to maintain spatial coherence in BEV. Given a rotation angle, the transformed semantic map is computed using a 2D rotation matrix:

\begin{equation} \begin{bmatrix} x' \\ y' \end{bmatrix}
\begin{bmatrix}
\cos\theta & -\sin\theta \\
\sin\theta & \cos\theta
\end{bmatrix}
\begin{bmatrix}
x - x_c \\ y - y_c
\end{bmatrix}
+ \begin{bmatrix}
x_c \\ y_c
\end{bmatrix}, \end{equation}

where 
 is the center of rotation.

In addition to flipping and rotation, BEV semantic map mixing is performed similarly to panoramic images:

\begin{equation} S' = \lambda S_1 + (1 - \lambda) S_2. \end{equation}

This ensures the semantic diversity of BEV annotations while preserving spatial alignment. The combination of these augmentation strategies enhances the robustness and generalization of the model in downstream tasks.


\clearpage

\end{document}